\documentclass[a4paper,fleqn]{cas-dc}

\usepackage[numbers]{natbib}

\def\tsc#1{\csdef{#1}{\textsc{\lowercase{#1}}\xspace}}
\tsc{WGM}
\tsc{QE}
\tsc{EP}
\tsc{PMS}
\tsc{BEC}
\tsc{DE}
\usepackage{amssymb}
\usepackage{algorithm}
\usepackage{dsfont}
\usepackage{algorithmic}
\usepackage{makecell}
\usepackage{hyperref}
\usepackage{etoolbox}

\definecolor{MyForestGreen}{rgb}{0.13, 0.55, 0.13} 

\begin{document}
\let\WriteBookmarks\relax
\def\floatpagepagefraction{1}
\def\textpagefraction{.001}
\shorttitle{}
\shortauthors{Qiwei Ma et~al.}

\title [mode = title]{Not All Patches are Equal: Sampling Matters for Visible-Infrared Pre-Training}                      



\author[1]{Qiwei Ma}[
orcid=0009-0000-2382-1416
]
\credit{Writing - Original Draft, Software, Methodology}

\author[1]{Bin Deng}
\credit{Conceptualization, Visualization}

\author[3]{Junjie Zhu}
\credit{Data curation}

\author[3]{Qiangjuan Huang}
\credit{Supervision}
\author[1]{Puhong Duan}
\credit{Writing - Review \& Editing}
\author[3]{Ke Yang}[
orcid=0000-0002-6923-8506
]
\cormark[1]
\credit{Writing - Review \& Editing, Methodology, Formal analysis, Validation}

\author[1,2]{Xudong Kang}
\credit{Project administration, Funding acquisition}

\author[1]{Shutao Li}
\credit{Project administration, Funding acquisition}

\affiliation[1]{organization={School of Artificial Intelligence and Robotics},
                addressline={Hunan University}, 
                city={Changsha},
                postcode={410082}, 
                state={Hunan},
                country={China}}
\affiliation[2]{organization={Yuelushan Center for Industrial Innovation},
                city={Changsha},
                postcode={410082}, 
                state={Hunan},
                country={China}}

\affiliation[3]{organization={Intelligent Game and Decision Lab},
                city={Beijing},
                postcode={100166}, 
                state={Beijing},
                country={China}}

\cortext[cor1]{Corresponding author}


\begin{abstract}
Visible-infrared (VIS-IR) alignment is a key pre-training task for robust multi-sensor perception. Most existing methods use uniform patch-wise contrastive learning, but this can be unreliable in VIS-IR data because imaging-physics differences make some spatially paired regions inherently less comparable, and aligning them with equal strength hinders representation learning and downstream transfer. In this paper, we revisit VIS-IR pre-training from a sampling perspective and propose Importance-Aware Sampling (IAS), which adjusts training emphasis based on patch reliability. Specifically, IAS (i) derives patch weights from infrared structural cues and uses them to reweight the contrastive objective; (ii) learns a soft importance mask with a lightweight sampler, optionally warm-started from the hand-crafted prior; and (iii) employs a patch curriculum learning strategy that gradually expands from high-reliability regions to harder patches. It is worth noting that IAS is plug-and-play and works with both patch-/correlation-level alignment (e.g., UNIV-style) and image-level contrastive baselines (e.g., ImageBind-style). Extensive experiments on multiple VIS-IR benchmarks demonstrate consistent improvements over strong baselines, including for IR semantic segmentation, IR object detection and VIS semantic segmentation and cross-modal retrieval task. Code will be released on \url{https://github.com/KlayMa527/IAS}.
\end{abstract}




\maketitle

\section{Introduction}
\label{sec:intro}

The combination of visible and infrared sensors is widely deployed in real-world multi-sensor perception systems, including night-time surveillance, autonomous driving, and multi-spectral remote sensing \cite{MSRS,CFD_DETR}. Learning aligned VIS-IR representations improves robustness under changing illumination and adverse conditions, making VIS-IR pre-training a common ingredient in recent multimodal pipelines \cite{Imagebind}. Along this line, a recent study investigates ViT-based patch-level alignment and cross-modal contrastive learning, for example, by aligning patch-to-patch similarity structures across modalities \cite{UNIV}.

\begin{figure}
	\centering
	\includegraphics[width=3.2in]{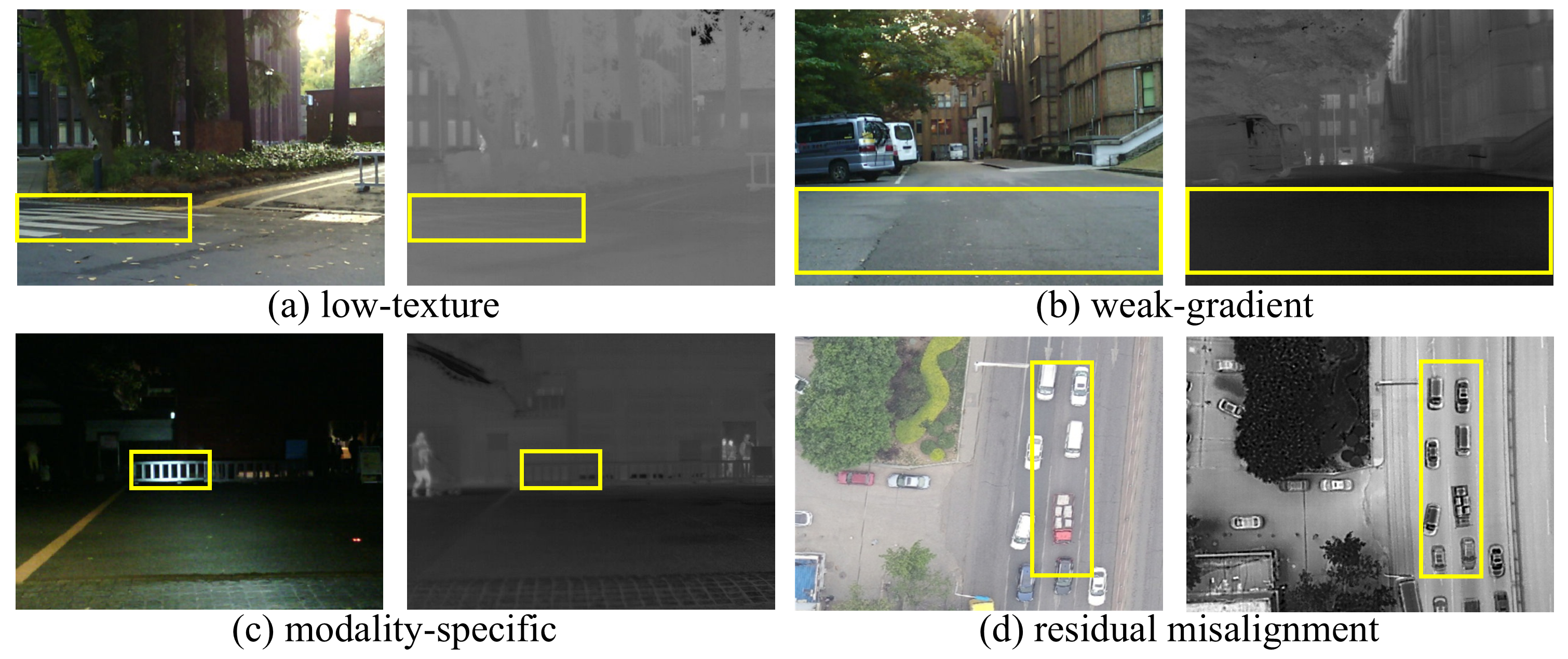}
	\caption{Factors that challenge VIS-IR alignment. We summarize four common sources of unreliable patch correspondence. (a) Low-texture regions in the infrared image provide limited discriminative detail. (b) Weak-gradient regions that are homogeneous in both visible and infrared (e.g., sky/road/walls) offer little structural cue for alignment. (c) Modality-specific effects may appear only in one modality (e.g., reflections, hot exhaust, sensor noise). (d) Residual misalignment remains after registration due to sensor geometry, parallax, and dynamics.}
	\label{fig_teaser}
\end{figure}

\begin{figure}
	\centering
	\includegraphics[width=3.2in]{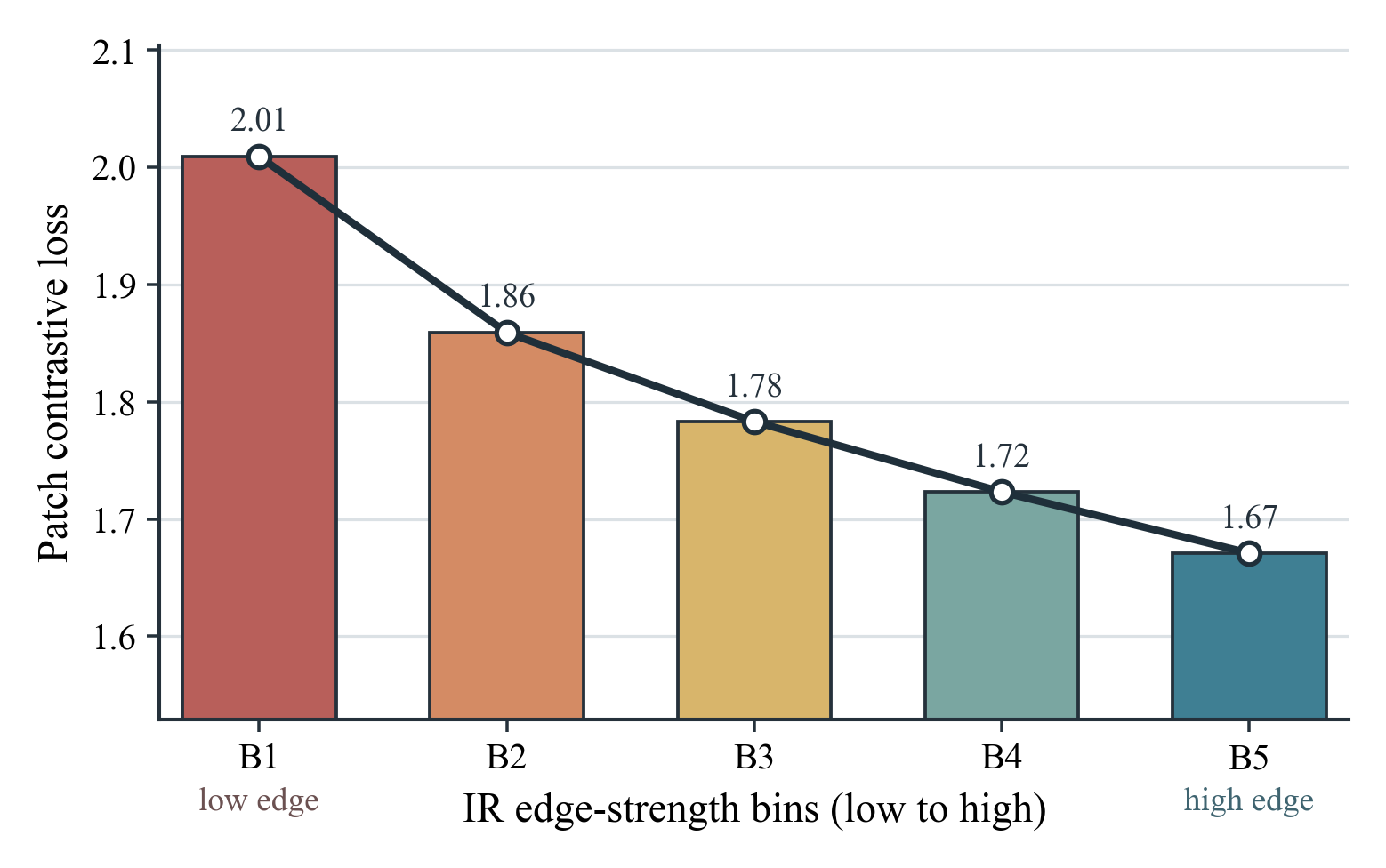}
	\caption{Patch-level contrastive loss decreases with increasing infrared edge strength.}
	\label{fig_bins}
\end{figure}

Despite encouraging progress, most VIS--IR alignment methods rely on a strong yet rarely examined assumption: given a registered VIS-IR pair, all spatially paired patches are equally informative and equally trustworthy. In a typical cross-modal contrastive learning pipeline, patches are sampled uniformly; co-located VIS and IR patches are treated as positives; negatives are drawn from other patches in the mini-batch; and the contrastive loss is averaged across patches with uniform weights \cite{UNIV,Imagebind}. However, this assumption is often violated in practice due to inherent sensing and imaging differences. As summarized in Fig.~\ref{fig_teaser}, patch correspondence can be unreliable because of (a) low-texture regions in infrared, (b) weak-gradient homogeneous areas in both modalities (e.g., sky/road/walls), (c) modality-specific effects that appear only in one modality, and (d) residual misalignment after registration caused by sensor geometry, parallax, and dynamics. Moreover, to better understand the source of cross-modal supervision, we analyze the relationship between infrared structural richness and patch-level contrastive difficulty. As shown in Fig. ~\ref{fig_bins}, infrared patches with lower edge strength consistently produce higher contrastive loss, while patches containing stronger edge responses are easier to align with visible-image representations. This observation suggests that textureless or weakly structured infrared regions provide ambiguous correspondence cues and may dominate the optimization with unreliable gradients. \textit{In this paper, we focus on the first three problems, under the assumption that pixel-level alignment has already been achieved.}

For such patches, enforcing equal-strength VIS-IR alignment is unnecessary and can even be harmful, since it introduces conflicting supervision and noisy gradients from regions that are only loosely matched across modalities. Empirically, uniformly aligning all patches tends to make optimization less stable and yields weaker downstream transfer. 
These observations lead to a simple but fundamental question: \textbf{Does sampling matter in VIS-IR alignment representation learning?} We answer this question affirmatively and argue that sampling and weighting should be treated as first-class design choices in VIS-IR pre-training. 
In this paper, we propose Importance-Aware Sampling (IAS), a lightweight and modular framework that estimates patch importance, reflecting how alignable a patch is across VIS and IR.
IAS treats sampling and weighting as part of the learning objective by incorporating patch weights as per-patch coefficients in the contrastive loss. As a simple yet effective instantiation, IAS starts from an infrared structural prior, where edge/gradient cues computed by Sobel or HOG correlate with regions that provide more reliable cross-modal correspondence. We compute patch-wise importance from these cues and use the resulting weights to emphasize structured and informative patches while down-weighting flat, noisy, or weakly matched regions.

Importantly, hand-crafted structural priors cannot capture all sources of unreliable correspondence, especially modality-specific effects that appear only in one modality (Fig.~\ref{fig_teaser}(c)). Such regions may still contain edges or gradients yet remain poorly alignable across modalities due to imaging-physics differences, and thus require adaptive identification beyond fixed cues. To this end, IAS further introduces a learnable sampling module that predicts a soft importance mask from infrared features in a self-supervised manner, optionally warm-started from the Sobel/HOG prior to stabilize early training. We further design an importance-curriculum learning schedule that gradually expands training from high-reliability patches to more ambiguous ones, improving optimization stability and strengthening alignment. In addition, IAS is plug-and-play, modifying only patch selection and loss weighting while leaving backbone architectures unchanged. It can be integrated into patch-/correlation-level alignment models (e.g., UNIV-style) and image-level contrastive baselines in an ImageBind-like setting via importance-aware pooling \cite{UNIV,Imagebind}. Our main contributions are summarized as follows:
\begin{itemize}
	\item We revisit VIS-IR pre-training from a sampling perspective and show that the widely used uniform patch sampling strategy is often sub-optimal, since patch pairs vary substantially in alignability due to imaging-physics differences.
	\item Building on this analysis, we propose Importance-Aware Sampling (IAS), a lightweight and plug-and-play sampling-centric framework that assigns patch importance to reweight patch-wise contrastive learning, instantiated with an infrared structural prior, an optional learnable sampler, and an importance curriculum learning schedule.
	\item Extensive experiments on multiple VIS-IR benchmarks show that IAS consistently improves segmentation and retrieval performance across alignment paradigms, supported by comprehensive ablations on sampling, weighting, and curriculum design.
\end{itemize}

\begin{figure*}
	\centering
	\includegraphics[width=6.8in]{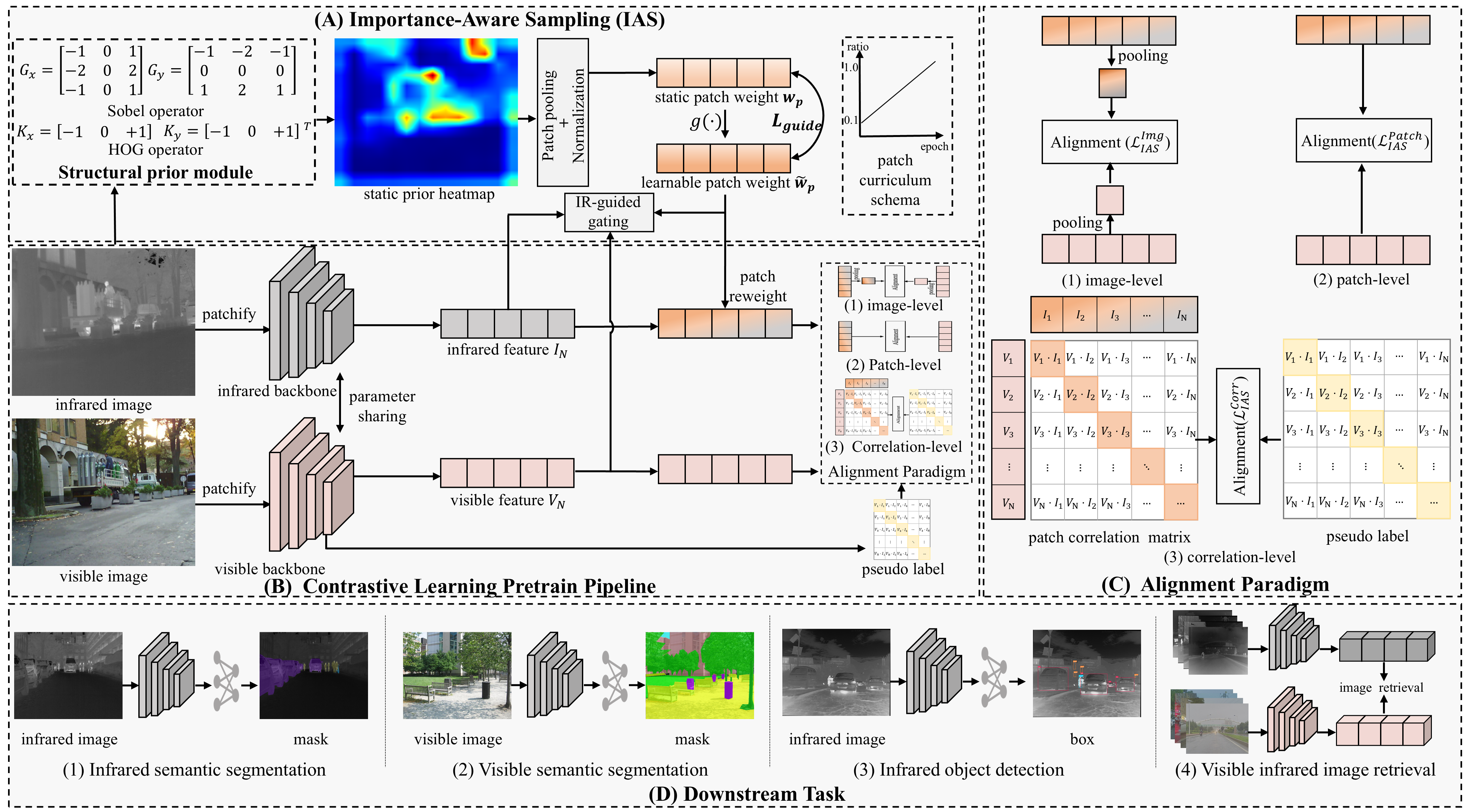}
	\caption{The overview of the proposal IAS framework. (A) Importance-aware sampling. (B) The Contrastive learning pretrain pipeline. (C) Alignment paradigm (C) The illustration of the downstream task. }
	\label{fig_framework}
\end{figure*}
\section{Related Work}

\subsection{Self-supervised Infrared Representation}
With the rapid progress of self-supervised learning, methods such as BYOL~\cite{BYOL} and MAE~\cite{MAE} have been extensively explored in visible-image scenarios. Meanwhile, driven by advances in sensing hardware and the increasing availability of infrared data, many recent studies have focused on foundation model research for the infrared modality in order to learn more general and transferable representations. INFMAE~\cite{INFMAE} constructs a large-scale infrared dataset and learns infrared representations using masked autoencoding, showing strong performance on downstream detection and segmentation tasks. Building on this line of work, PAD~\cite{PAD} and UNIP~\cite{UNIP} further scale up and diversify single-modality infrared pre-training data, and they incorporate distillation-based learning to achieve additional performance gains. 

Despite their promising results, these infrared foundation models remain limited to the infrared modality, which prevents them from learning unified cross-modal representations between visible and infrared and also constrains them by the inherent limitations of single-modality information. In this paper, we systematically investigate contrastive learning paradigms for visible–infrared pre-training and discuss them in conjunction with our IAS strategy.

\subsection{Visible-Infrared Representation Learning}
Visible-infrared representation learning has been increasingly utilized for multispectral detection~\cite{GLFNet,Dronevehicle,Lrafnet,COMO}, multispectral segmentation~\cite{TINN,Bridging}, cross-modal retrieval~\cite{liu2021infrared}, cross-modal re-identification~\cite{park2021learning}, and image fusion~\cite{mo2021attribute, SPGFusion,Interactive}. Early approaches were mainly built upon hand-crafted descriptors with explicit matching~\cite{YE2019CFOG}. Recently, cross-modal contrastive learning methods such as CLIP~\cite{CLIP,SARVLM} have advanced significantly. As a result, there has been increasing interest in learning cross-modal representations between visible and infrared modalities, where contrastive objectives are used to align the two modalities in a shared embedding space. ImageBind~\cite{Imagebind} learns a unified embedding over six modalities, including visible and infrared signals. Beyond global alignment, patch-level correspondence has been explored for fine-grained visible-infrared alignment. UNIV~\cite{UNIV} adopts a patch correlation alignment framework built on ViT~\cite{MCMAE}, where patch-wise similarity structures are aligned across modalities and a frozen RGB encoder is commonly employed to provide stable pseudo supervision. 

Despite their effectiveness, these methods typically assume that all patches in an aligned pair contribute equally to alignment. Although attention-based heuristics have been used to suppress noisy correspondences, patch-wise importance from the infrared modality is rarely modeled explicitly, and sampling is seldom treated as a dedicated learning component.

\section{Method}
\label{sec:method}

\subsection{Overview}
\label{sec:method_overview}

We study visible-infrared representation learning with pixel-aligned image pairs. The key idea of IAS is to explicitly estimate patch importance from infrared structural cues and incorporate such importance into visible-infrared alignment. As shown in Fig.~\ref{fig_framework}, IAS first derives per-patch importance weights from infrared images, and then uses them to reweight image-level, patch-level, or correlation-based alignment objectives. In addition, the hand-crafted structural prior can be replaced by a lightweight learnable sampler, which is initialized by Sobel-guided warm-up and optimized with an importance curriculum. IAS is plug-and-play and does not modify the backbone encoder.

\subsection{Problem Setup}

We consider a dataset of pixel-aligned visible-infrared image pairs
\(\mathcal{D}=\{(\mathbf{x}_b^V,\mathbf{x}_b^I)\}_{b=1}^{B}\), where \(\mathbf{x}_b^V\) and \(\mathbf{x}_b^I\) denote the visible image and the corresponding infrared image in a mini-batch. Each image is divided into \(N\) non-overlapping patches. The visible and infrared branches, denoted by \(f_V\) and \(f_I\), encode the two modalities into patch embeddings:
\begin{equation}
	\mathbf{Z}_b^V=\{\mathbf{z}_{b,p}^V\}_{p=1}^{N}=f_V(\mathbf{x}_b^V),
\end{equation}
\begin{equation}
	\mathbf{Z}_b^I=\{\mathbf{z}_{b,p}^I\}_{p=1}^{N}=f_I(\mathbf{x}_b^I),
\end{equation}
where \(\mathbf{z}_{b,p}^V,\mathbf{z}_{b,p}^I\in\mathbb{R}^{d}\) denote the \(d\)-dimensional embeddings of the \(p\)-th patch. For image-level alignment, the patch embeddings are further aggregated into global representations:
\begin{equation}
	\mathbf{h}_b^V=\phi_V(\mathbf{Z}_b^V),
\end{equation}
\begin{equation}
	\mathbf{h}_b^I=\phi_I(\mathbf{Z}_b^I),
\end{equation}
where \(\phi_V(\cdot)\) and \(\phi_I(\cdot)\) denote the projection layers.

\begin{table}[t]
	\centering
    \scriptsize
	\renewcommand{\arraystretch}{1.05} 
	\caption{Illustration of pre-training datasets.}
	\label{tab_dataset}
	\setlength{\tabcolsep}{0.1pt}
	\begin{tabular*}{\hsize}{@{\extracolsep{\fill}}lcccccc}
		\toprule
		dataset   & Image Count & Pre-training Paradigm & Model & Infrared      & Visible  & Paired \\ \midrule
		MSIP      & 178,756    & Masked Modeling      & PAD      & $\checkmark$   &       &\\
		INF30     & 305,241    & Masked Modeling      & INFMAE   & $\checkmark$   &       & \\
		INFMIX    & 859,375    & Masked Modeling      & UNIP     & $\checkmark$     & $\checkmark$  & \\
		MVIP      & 98,992     & Contrastive Learning & UNIV,IAS & $\checkmark$  & $\checkmark$  & $\checkmark$ \\
		\bottomrule
	\end{tabular*}	
\end{table}

\noindent\textbf{Image-level Visible-Infrared Alignment.}
For image-level contrastive learning, each paired visible-infrared image forms a positive pair \((\mathbf{h}_b^V,\mathbf{h}_b^I)\), while infrared images from other samples in the mini-batch are treated as negatives. The image-level contrastive loss is defined as
\begin{equation}
	\ell_{\mathrm{img}}(b)=
	-\log
	\frac{
		\exp(\mathrm{sim}(\mathbf{h}_b^V,\mathbf{h}_b^I)/\tau_g)
	}{
		\sum_{r=1}^{B}
		\exp(\mathrm{sim}(\mathbf{h}_b^V,\mathbf{h}_r^I)/\tau_g)
	},
	\label{eq:image_infonce}
\end{equation}
where \(\mathrm{sim}(\cdot,\cdot)\) denotes cosine similarity and \(\tau_g\) is the image-level temperature. The reverse infrared-to-visible direction can be symmetrically included.

\noindent\textbf{Patch-level Visible-Infrared Alignment.}
At the patch level, patches at the same spatial location are regarded as positive pairs, while infrared patches from different locations or different images serve as negatives. For the \(p\)-th visible patch of the \(b\)-th image, the patch-level contrastive loss is formulated as
\begin{equation}
	\ell_{\mathrm{pair}}(b,p)=
	-\log
	\frac{
		\exp(\mathrm{sim}(\mathbf{z}_{b,p}^V,\mathbf{z}_{b,p}^I)/\tau_p)
	}{
		\sum_{(r,q)\in\mathcal{Q}}
		\exp(\mathrm{sim}(\mathbf{z}_{b,p}^V,\mathbf{z}_{r,q}^I)/\tau_p)
	},
	\label{eq:pair_infonce}
\end{equation}
where \(\tau_p\) is the patch-level temperature and \(\mathcal{Q}\) denotes the candidate set of infrared patch tokens, e.g., all infrared patches in the current mini-batch.

\noindent\textbf{Correlation-based Alignment.}
Following the UNIV-style formulation~\cite{UNIV}, we also consider patch-to-patch correlation alignment. Specifically, a similarity matrix \(\mathbf{S}_b^{IV}\in\mathbb{R}^{N\times N}\) is constructed between infrared and visible patch embeddings:
\begin{equation}
	\mathbf{S}_{b,ij}^{IV}=\mathrm{sim}(\mathbf{z}_{b,i}^I,\mathbf{z}_{b,j}^V),
	\label{eq:similarity_matrix}
\end{equation}
where \(\mathbf{S}_{b,ij}^{IV}\) measures the similarity between the \(i\)-th infrared patch and the \(j\)-th visible patch. This matrix is aligned with a pseudo-label matrix \(\mathbf{Y}_b\in\{0,1\}^{N\times N}\) derived from the last-layer visible attention map. The correlation-based alignment loss is defined as
\begin{equation}
	\mathcal{L}_{\mathrm{corr}}=
	\frac{1}{B N^2}
	\sum_{b=1}^{B}
	\sum_{i=1}^{N}
	\sum_{j=1}^{N}
	\mathrm{BCE}\left(
	\sigma(\mathbf{S}_{b,ij}^{IV}),
	\mathbf{Y}_{b,ij}
	\right),
	\label{eq:corr_loss}
\end{equation}
where \(\sigma(\cdot)\) denotes the sigmoid function and \(\mathrm{BCE}(\cdot,\cdot)\) denotes the binary cross-entropy loss.

\subsection{Structural Prior for Patch Importance}
\label{sec:structural_prior}

Uniformly treating all patches as equally informative is suboptimal for visible-infrared alignment. Low-texture background regions usually provide weak or noisy supervision, whereas structured regions containing edges, object boundaries, and distinctive textures tend to offer more reliable cross-modal correspondence. Therefore, IAS assigns each infrared patch an importance weight to measure its reliability for alignment.

\noindent\textbf{Sobel/HOG-based Structural Strength.}
Infrared images preserve structural information under challenging illumination. We compute a structural response map \(\mathbf{E}_b\in\mathbb{R}^{H\times W}\) from the infrared image \(\mathbf{x}_b^I\) using Sobel operators or HOG responses. Following the same patch partition as the vision transformer, the structural strength of the \(p\)-th patch is obtained by
\begin{equation}
	s_{b,p}=\mathrm{Pool}\left(\mathbf{E}_{b,\mathcal{P}(p)}\right),
	\label{eq:sp_pool}
\end{equation}
where \(\mathcal{P}(p)\) denotes the spatial region of the \(p\)-th patch, and \(\mathrm{Pool}(\cdot)\) is average pooling. The patch importance weight is then normalized as
\begin{equation}
	w_{b,p}=
	\frac{s_{b,p}}
	{\max_{q\in\{1,\dots,N\}} s_{b,q}+\varepsilon},
	\label{eq:wp_norm}
\end{equation}
where \(\varepsilon\) is a small constant for numerical stability. This normalization constrains \(w_{b,p}\in[0,1]\), where a larger value indicates stronger infrared structural response.

\subsection{Importance-Aware Alignment Objectives}
\label{sec:ias_objectives}

Given the patch importance weights \(\{w_{b,p}\}_{p=1}^{N}\), IAS reweights the contribution of patches or patch pairs in different visible-infrared alignment objectives. Let \(\omega_{b,p}\) denote the final importance weight used for training, which can be instantiated by the hand-crafted prior \(w_{b,p}\) or the learnable weight \(\tilde{w}_{b,p}\) introduced in Sec.~\ref{sec:learnable_sampler}.

\noindent\textbf{Image-level Objective.}
For image-level contrastive baselines, IAS constructs an importance-pooled infrared representation:
\begin{equation}
	\hat{\mathbf{h}}_b^I
	=
	\sum_{p=1}^{N}
	\alpha_{b,p}\mathbf{z}_{b,p}^I,
	\quad
	\alpha_{b,p}
	=
	\frac{\omega_{b,p}}
	{\sum_{q=1}^{N}\omega_{b,q}+\varepsilon}.
	\label{eq:global_pool}
\end{equation}
The image-level contrastive loss in Eq.~\eqref{eq:image_infonce} is then applied to \((\mathbf{h}_b^V,\hat{\mathbf{h}}_b^I)\), enabling global visible-infrared alignment to focus more on structurally informative infrared regions.

\noindent\textbf{Patch-level Objective.}
For patch-level contrastive learning, IAS weights the loss of each visible-infrared patch pair:
\begin{equation}
	\mathcal{L}_{\mathrm{IAS}}^{\mathrm{patch}}
	=
	\frac{1}{B}
	\sum_{b=1}^{B}
	\frac{1}{Z_b}
	\sum_{p=1}^{N}
	\omega_{b,p}
	\ell_{\mathrm{pair}}(b,p),
	\label{eq:ias_patch}
\end{equation}
where \(Z_b=\sum_{p=1}^{N}\omega_{b,p}+\varepsilon\) normalizes the loss scale. This formulation preserves the original contrastive structure while reducing the contribution of low-importance infrared regions.

\noindent\textbf{Correlation-based Objective.}
For UNIV-style correlation alignment, IAS reweights each patch-to-patch correlation according to the importance of the corresponding spatial locations:
\begin{equation}
	\mathcal{L}_{\mathrm{IAS}}^{\mathrm{corr}}
	=
	\frac{1}{B}
	\sum_{b=1}^{B}
	\frac{1}{Z_b^{\mathrm{corr}}}
	\sum_{i=1}^{N}
	\sum_{j=1}^{N}
	\omega_{b,i}\omega_{b,j}
	\mathrm{BCE}\left(
	\sigma(\mathbf{S}_{b,ij}^{IV}),
	\mathbf{Y}_{b,ij}
	\right),
	\label{eq:ias_corr}
\end{equation}
where \(Z_b^{\mathrm{corr}}=\sum_{i=1}^{N}\sum_{j=1}^{N}\omega_{b,i}\omega_{b,j}+\varepsilon\). Eq.~\eqref{eq:ias_corr} suppresses unreliable patch correlations while retaining strong supervision from structurally informative regions.

\subsection{Learnable Sampling Module}
\label{sec:learnable_sampler}

Although Sobel/HOG priors are simple and effective, they mainly capture low-level structural responses. To adaptively model higher-order semantic cues, we introduce a lightweight learnable sampler that predicts patch importance from infrared patch embeddings.

\noindent\textbf{Importance Prediction.}
Given infrared patch embeddings \(\mathbf{Z}_b^I\), a lightweight network \(g_{\theta}\) predicts per-patch logits:
\begin{equation}
	\mathbf{m}_b=g_{\theta}(\mathbf{Z}_b^I),
	\quad
	\mathbf{m}_b\in\mathbb{R}^{N}.
	\label{eq:logits}
\end{equation}
The logits are converted into soft importance weights by a temperature-controlled sigmoid:
\begin{equation}
	\tilde{w}_{b,p}
	=
	\sigma\left(\frac{\mathbf{m}_{b,p}}{T_s}\right),
	\label{eq:soft_weight}
\end{equation}
where \(T_s>0\) controls the sharpness of the predicted weights. A smaller \(T_s\) produces more selective patch weights.

\noindent\textbf{Sobel-guided Warm-up.}
Directly learning patch importance from scratch can be unstable at early training stages. Therefore, the Sobel-based prior \(w_{b,p}^{\mathrm{Sobel}}\) computed by Eq.~\eqref{eq:sp_pool} and Eq.~\eqref{eq:wp_norm} is used as a teacher signal during warm-up:
\begin{equation}
	\mathcal{L}_{\mathrm{guide}}
	=
	\frac{1}{B N}
	\sum_{b=1}^{B}
	\sum_{p=1}^{N}
	\left|
	\tilde{w}_{b,p}
	-
	w_{b,p}^{\mathrm{Sobel}}
	\right|.
	\label{eq:guide}
\end{equation}
The training objective during warm-up is
\begin{equation}
	\mathcal{L}
	=
	\mathcal{L}_{\mathrm{align}}^{\mathrm{IAS}}(\tilde{w})
	+
	\lambda_{\mathrm{guide}}
	\mathcal{L}_{\mathrm{guide}},
	\label{eq:total_loss}
\end{equation}
where \(\mathcal{L}_{\mathrm{align}}^{\mathrm{IAS}}\) denotes the selected IAS objective in Eq.~\eqref{eq:ias_patch} or Eq.~\eqref{eq:ias_corr}, and \(\lambda_{\mathrm{guide}}\) controls the strength of Sobel guidance. After warm-up, \(\lambda_{\mathrm{guide}}\) is decayed to zero, allowing the sampler to refine patch importance in a data-adaptive manner.

\subsection{Patch Curriculum over Importance}
\label{sec:curriculum}

To stabilize early-stage optimization, IAS adopts a curriculum strategy that gradually increases the fraction of patches participating in the alignment loss. At epoch \(t\), the coverage ratio is defined as
\begin{equation}
	\rho(t)
	=
	\rho_{\min}
	+
	(\rho_{\max}-\rho_{\min})
	\min\left(\frac{t}{T},1\right),
	\label{eq:rho}
\end{equation}
where \(0<\rho_{\min}\leq\rho_{\max}\leq1\), and \(T\) denotes the curriculum length. For each image, patches are sorted according to their importance scores. The curriculum mask is defined as
\begin{equation}
	c_{b,p}(t)=
	\begin{cases}
		1, & \mathrm{rank}(\omega_{b,p}) \leq \lceil \rho(t)N\rceil, \\
		\eta, & \mathrm{otherwise},
	\end{cases}
	\label{eq:curriculum_mask}
\end{equation}
where \(\eta\in[0,1)\) is a small retaining factor. The final training weight is then given by
\begin{equation}
	\bar{\omega}_{b,p}(t)=\omega_{b,p}c_{b,p}(t).
	\label{eq:final_weight}
\end{equation}
In practice, \(\bar{\omega}_{b,p}(t)\) replaces \(\omega_{b,p}\) in Eq.~\eqref{eq:ias_patch} or Eq.~\eqref{eq:ias_corr}. This curriculum encourages the model to first learn from reliable structured regions and gradually incorporate more ambiguous patches as training progresses.

\subsection{Integration into Existing Frameworks}
\label{sec:integration}
\begin{table}[t]
	\centering
    \scriptsize
    \renewcommand{\arraystretch}{1.05}
	\setlength{\tabcolsep}{0pt}
	\caption{Experiment result on infrared semantic segmentation task. $^{\dagger}$ indicate results from our reproduce checkpoint on official code while $^{\ast}$ denotes results adapted form official checkpoint.}
	\begin{tabular*}{\hsize}{@{\extracolsep{\fill}}lcccccc}
		\toprule
		\multirow{2}{*}{Method} & Pretrained & Pretrained  & MFNet & SODA  & SCUTSEG & MSRS\\ \cline{4-7}
		& Data & Epoch & mIoU   & mIoU   & mIoU   & mIoU  \\
		\midrule
		DeepLab V3+         & -- & -- & 49.80 & 68.73 & 50.46 & 65.20   \\
		PSPNet              & -- & -- & 45.24 & 68.68 & 48.16 & --  \\
		UPerNet             & -- & -- & 48.56 & 67.45 & -- & 65.60 \\
		SegFormer           & -- & -- & 50.68 & 67.86 & -- & --  \\
		ViT-adapter         & -- & -- & 50.62 & 68.12 & -- & --  \\
		Mask2Former         & -- & -- & 51.30 & 67.58 & -- & --  \\
		MaskDINO            & -- & -- & 51.03 & 66.32 & -- & --  \\
		EC-CNN              & -- & -- & 47.56 & 65.87 & -- & --  \\
		TINN                & -- & -- & 43.15 & 69.45 & -- & --  \\
		\midrule
		INFMAE $^{\ast}$  & INF30    & 400   & 44.78 & 60.26 & 64.33  & 73.14\\
		MCMAE               & IN1K   & 100 & 45.36 & 64.90 & -- & 74.90  \\
		PAD                 & MSIP   & 100 & 48.82 & 68.41 & -- & --  \\
		UNIV(LoRA)          & MSIP   & 100 & 51.06 & 69.60 & -- & --  \\
		\midrule
		UNIV $^{\ast}$      & MVIP   & 400 & 50.64 & 70.22 & 68.49   & 75.54   \\
		UNIV $^{\dagger}$   & MVIP   & 400 & 51.14 & 69.68 & 69.65   & 75.48   \\
		IAS(Image-style)    & MVIP   & 400 & 52.27 & \textbf{71.07}  & 71.02 & 75.65   \\
		IAS(UNIV-style)     & MVIP   & 400 & \textbf{52.32} & 70.59  & \textbf{72.23} & \textbf{75.85}  \\
		\multicolumn{3}{l}{IAS (UNIV-style) vs. UNIV $^{\ast}$}    & \textcolor{MyForestGreen}{+1.68}  & \textcolor{MyForestGreen}{+0.37}  & \textcolor{MyForestGreen}{+3.74}   & \textcolor{MyForestGreen}{+0.31}  \\
		\multicolumn{3}{l}{IAS (UNIV-style) vs. UNIV $^{\dagger}$}    & \textcolor{MyForestGreen}{+1.18}  & \textcolor{MyForestGreen}{+0.91}  & \textcolor{MyForestGreen}{+2.58}   & \textcolor{MyForestGreen}{+0.37}  \\
		\bottomrule
	\end{tabular*}
	\label{table_segmentaton}
\end{table}
IAS can be integrated into existing visible-infrared pretraining frameworks without changing their encoder architectures. For image-level contrastive baselines, IAS replaces the original infrared global representation with the importance-pooled representation in Eq.~\eqref{eq:global_pool}. For patch-level contrastive frameworks, IAS reweights the patch-level contrastive loss by Eq.~\eqref{eq:ias_patch}. For UNIV-style correlation alignment, IAS applies importance-aware correlation weighting by Eq.~\eqref{eq:ias_corr}. This gating weakens supervision from infrared-unreliable regions while preserving confident structural correspondences. Therefore, IAS only modifies loss construction or global pooling, making it a low-cost and plug-and-play module for visible-infrared representation learning.
\section{Experiments}
\begin{table}[t]
	\centering
    \renewcommand{\arraystretch}{1.05}
	\footnotesize
	\caption{Experiment result on visible-infrared retrieval task. $^{\dagger}$ and $^{\ast}$ are the same meaning as the previously settings. }
	\setlength{\tabcolsep}{0.5pt}
	\begin{tabular*}{\hsize}{@{\extracolsep{\fill}}lccccccc}
		\toprule
		Method & Pretrained  & \multicolumn{3}{c}{Visible2Infrared} & \multicolumn{3}{c}{Infrared2Visible}  \\ \cline{3-8}
		&  Data       & R@1    & R@5   & R@10 & R@1 & R@5  & R@10  \\
		\midrule
		MCMAE               & IN1K   & 1.35 & 3.67 & 5.67 & 1.19 & 3.31 & 5.12  \\
		INFMAE              & INF30  & 0.95 & 3.09 & 5.00 & 1.19 & 3.80 & 7.02 \\
		UNIP                & INFMIX & 0.24 & 0.60 & 0.83 & 0.12 & 0.60 & 1.19 \\
		\midrule
		UNIV $^{\ast}$         & MVIP	 & 15.95 & 37.50 & 48.69 & 12.50 & 29.76 & 39.52\\
		UNIV $^{\dagger}$      & MVIP  & 12.38 & 31.19 & 43.21 & 9.88 & 22.50 & 30.47 \\
		IAS(UNIV-style)        & MVIP   & 13.80 & 33.21 & 44.52 & 10.23 & 23.57 & 29.64 \\
		IAS(Image-style)       & MVIP   & \textbf{82.14} & \textbf{95.35} & \textbf{98.45} & \textbf{83.33} & \textbf{95.23} & \textbf{98.21} \\
		\bottomrule
	\end{tabular*}
	\label{table_retrieval}
\end{table}
\subsection{Experiment setting}
\noindent\textbf{Dataset.} For pre-training, we use the MVIP dataset from~\cite{UNIV}, which contains 98,992 aligned visible-infrared image pairs. The INF30~\cite{INFMAE}, MSIP~\cite{UNIV}, and INFMIX~\cite{UNIP} contain 178,756, 305,241, and 859,375 single modality images, respectively. More details of pre-training datasets are described in Table~\ref {tab_dataset}. To evaluate the effectiveness of our IAS strategy, we conduct experiments on infrared object segmentation, infrared object detection, visible segmentation, and visible-infrared retrieval, which are introduced as follows:

\noindent\textbf{MFNet}~\cite{MFNet} is a visible-infrared semantic segmentation dataset for urban driving scenes. It contains 1,569 aligned RGB-T image pairs with pixel-level annotations for eight obstacle classes plus background. We use 784, 392, and 393 samples for training, validation, and testing, respectively. 

\noindent\textbf{MSRS}~\cite{MSRS} is a high-quality aligned visible-infrared dataset for road-scene understanding. It comprises 1,444 pixel-level annotated image pairs with nine semantic categories. We utilize 1,083 pairs for training and 361 pairs for testing. 

\noindent\textbf{SODA}~\cite{SODA} is a thermal infrared semantic segmentation dataset covering diverse indoor and outdoor scenes. It provides pixel-level annotations for 20 semantic categories, with 1,168 images for training and 1,000 images for testing.  

\noindent\textbf{SCUTSEG}~\cite{SCUT-SEG} contains 1,345 training images and 665 test images collected from nighttime driving scenes, with 10 semantic classes including road, person, and others. 
\begin{figure*}[!tbp]
	\centering
	\includegraphics[width=6.8in]{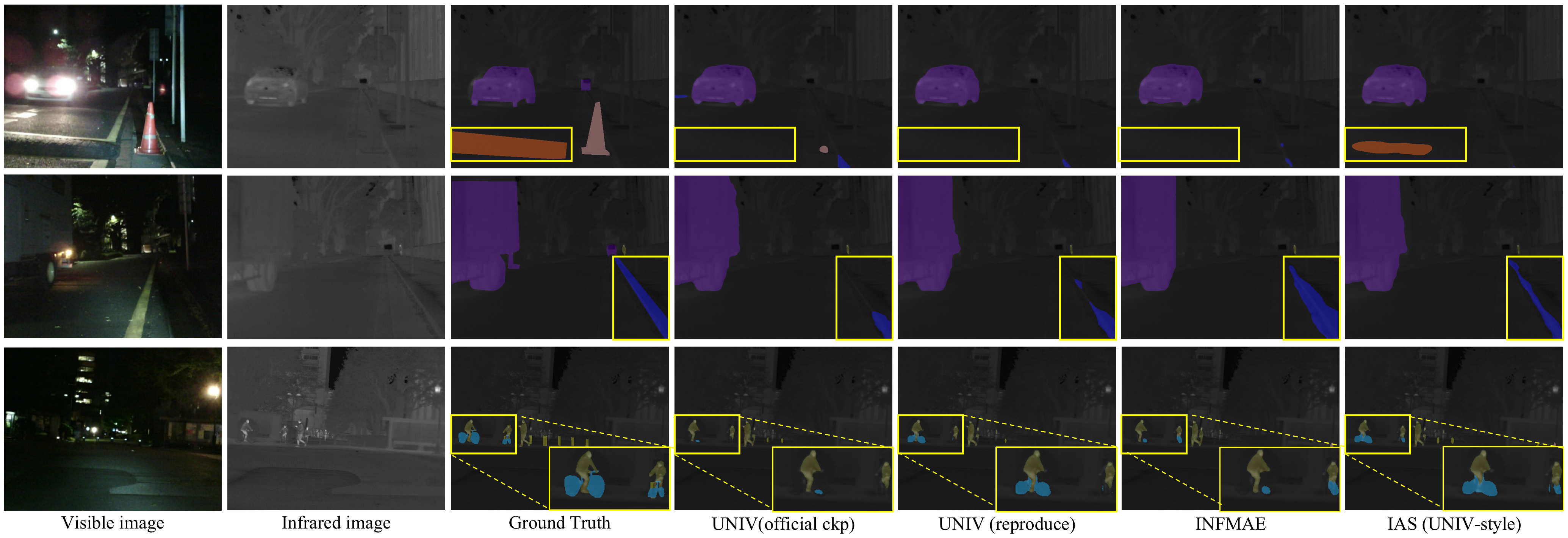}
	\caption{The infrared semantic segmentation experiment result on the MFNet dataset.}
	\label{fig_segmentaton_result_mfnet}
\end{figure*}
\begin{figure*}[!tbp]
	\centering
	\includegraphics[width=6.8in]{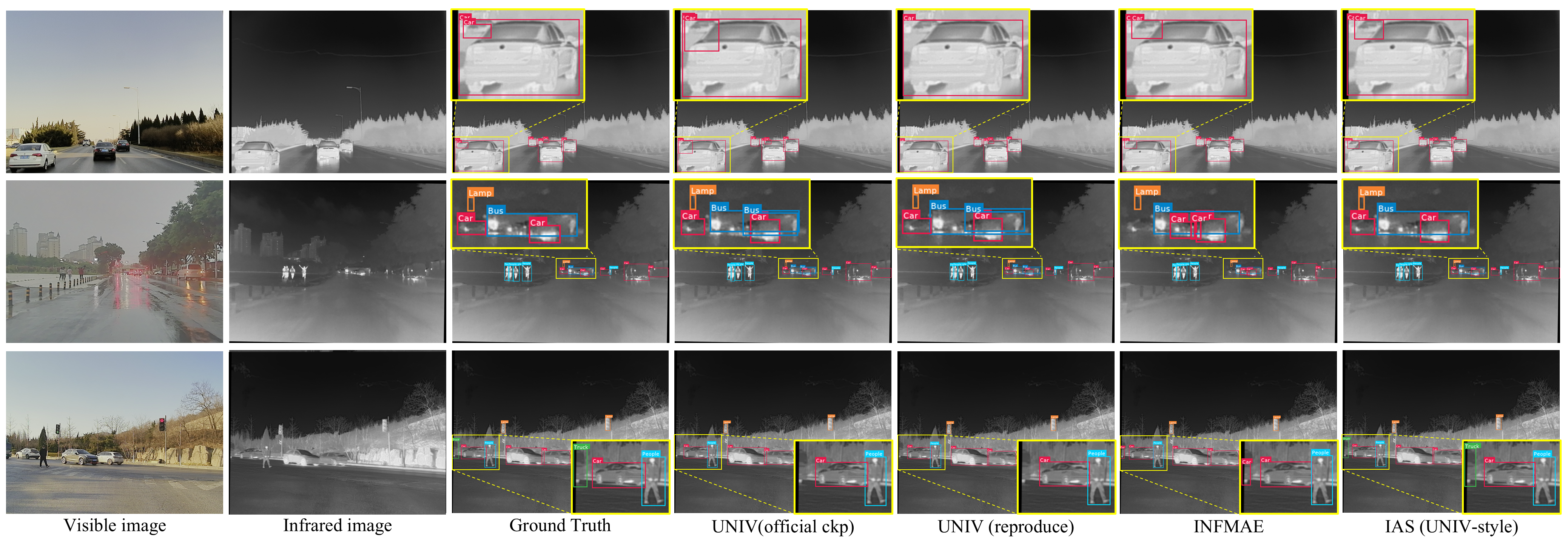}
	\caption{The infrared object detection experiment result on the M3FD dataset.}
	\label{fig_detection_result_m3fd}
\end{figure*}

\noindent\textbf{ADE20K}~\cite{ade20k} is a widely used visible semantic segmentation dataset, containing 20,210 training images and 2,000 validation images across diverse scenes, ranging from indoor and outdoor environments to natural and urban scenarios. It provides dense pixel-level annotations for 150 semantic categories.

\noindent\textbf{M3FD}~\cite{M3FD} is a paired visible-infrared dataset collected from diverse driving scenarios. Following~\cite{EAEFNet}, it contains 3,360 training pairs and 840 test pairs. We use the infrared images for object detection under the standard setting. For visible-infrared retrieval, we further sample 840 paired images from the original 4,200 image pairs to conduct zero-shot retrieval evaluation.

\noindent\textbf{Evaluation Metric.} 
The segmentation performance is evaluated using mean Intersection over Union (mIoU), mean class accuracy (mAcc), and overall pixel accuracy (aAcc). 
The detection performance is evaluated at mAP, AP50, and AP75 metrics.
The retrieval performance is evaluated using Recall at Top-$K$ (R@K), where $K \in \{1, 5, 10\}$. We report both infrared-to-visible and visible-to-infrared retrieval results, and compute mRecall(in Table~\ref{tab_abla}) as the average of the six R@K values:
\begin{equation}
	\mathrm{mRecall} =
	\frac{1}{6}
	\sum_{K \in \{1,5,10\}}
	\left(
	R@K_{\mathrm{I2V}} + R@K_{\mathrm{V2I}}
	\right)
\end{equation}

\noindent\textbf{Implementation Details.}
The learnable sampling module was implemented as a two-layer MLP with hidden dimension 256 and GELU activation, and was applied to each infrared patch embedding. The sampling temperature was set to $T_s=0.5$. The guidance weight $\lambda_{\text{guide}}$ was linearly decayed from 1.0 to 0.2 over the first 5 epochs. The curriculum parameters were set to $\rho_{\min}=0.2$ and $\rho_{\max}=1.0$.

A ViT-B/16 backbone from~\cite{MCMAE} was adopted with patch size $16\times16$ and embedding dimension $d=768$. For UNIV-style patch correlation alignment, the original architecture and hyperparameters were followed, and IAS was introduced only in the training loss. Optimization was performed using AdamW with an initial learning rate of $1\mathrm{e}{-4}$, weight decay of $0.05$, and batch size of 84, and training was conducted for 400 epochs.

For the static sampling strategy, the Sobel-based prior was computed by applying a $3\times3$ Sobel operator to infrared images to obtain the gradient magnitude, followed by average pooling within each patch to produce patch-level scores. For the HOG-based prior, 8 orientation bins were used, and the cell size was set to one quarter of the patch size. All edge scores were normalized per image.

The detailed setting of downstream segmentation and detection tasks are illustrated in Table~\ref{tab_downstream_settings}. 

\begin{table}[t]
	\centering
	\renewcommand{\arraystretch}{1.05}
	\caption{Experiment results on Infrared detection. $^{\dagger}$ and $^{\ast}$ are the same meaning as the previously settings.}
	\setlength{\tabcolsep}{0.1pt}
	\begin{tabular*}{\hsize}{@{\extracolsep{\fill}}lcccc}
		\toprule
		& \multicolumn{4}{c}{Infrared Detection (M3FD)} \\
		\cmidrule(r){2-5} 
		Method            & Modality & mAP   & AP50  & AP75    \\ \midrule
		DAMSDet           & IR+VIS & 52.90  & 80.2   & 56.0 \\       
		CDDFuse           & IR+VIS & 53.60  & 80.5   & -- \\  
		EMMA              & IR+VIS & 55.40  & 82.9   & -- \\ \midrule
		DAMSDet-DINO      & IR & 35.00  & 58.80   & 36.10 \\
		YOLOv8l           & IR & 53.10  & 79.50   & --  \\ \midrule
		MCMAE             & IR & 55.31  & 87.99  & 59.15   \\
		INFMAE            & IR & 55.81  & 87.67   & 59.36   \\
		UNIV$^\ast$       & IR & 55.17  & 87.24   & 58.65   \\
		UNIV$^\dagger$    & IR & 55.43  & 87.52   & 59.40   \\
		IAS (Image-style) & IR & 55.50   & \textbf{88.33}   & 58.61 \\
		IAS (UNIV-style)  & IR & \textbf{56.17}  & 87.89   & \textbf{60.83}  \\
		\multicolumn{2}{l}{IAS (UNIV-style) vs. UNIV$^\ast$} & \textcolor{MyForestGreen}{+1.00} & \textcolor{MyForestGreen}{+0.65} & \textcolor{MyForestGreen}{+2.18} \\
		\multicolumn{2}{l}{IAS (UNIV-style) vs. UNIV$^{\dagger}$} & \textcolor{MyForestGreen}{+0.74} & \textcolor{MyForestGreen}{+0.37} & \textcolor{MyForestGreen}{+1.43} \\
		\bottomrule
	\end{tabular*}
	\label{tab_m3fd}
\end{table}

\subsection{Comparison with State-of-the-Art Methods}
We compare the proposed IAS with representative methods on three downstream tasks, including infrared semantic segmentation, infrared object detection, and visible semantic segmentation. For semantic segmentation, the compared methods include classical segmentation frameworks, transformer-based models, mask-based architectures, and infrared-oriented methods, such as DeepLab V3+~\cite{Deeplabv3+}, PSPNet~\cite{PSPNet}, UPerNet~\cite{UperNet}, SegFormer~\cite{SegFormer}, ViT-Adapter~\cite{VIT-adapter}, Mask2Former~\cite{Mask2former}, MaskDINO~\cite{MaskDINO}, EC-CNN~\cite{EC-CNN}, and TINN~\cite{TINN}. We also compare IAS with well-pretrained backbones, including INFMAE~\cite{INFMAE}, MCMAE~\cite{MCMAE}, PAD~\cite{PAD}, and UNIV~\cite{UNIV}, all of which are equipped with the same UPerNet head for fair comparison. For infrared object detection, IAS is evaluated with a Mask R-CNN~\cite{maskrcnn} head and compared with representative detection and fusion-based methods, including DAMSDet~\cite{damsdet}, EMMA~\cite{EMMA}, CDDFuse~\cite{CDDFuse}, YOLOv8~\cite{yolov8}, and other pretrained backbones. For visible semantic segmentation, we further compare IAS with classical segmentation methods and recent pretrained models, including C-JEPA~\cite{C-JEPA}, GATE~\cite{GATE}, PE Rescale~\cite{perescale}, CR2PQ~\cite{cr2pq}, and ViM-VQ~\cite{vim}.

\begin{table}[t]
	\centering
	\renewcommand{\arraystretch}{1.05}
	\caption{Experiment results on visible segmentation. $^{\dagger}$ and $^{\ast}$ are the same meaning as the previously settings.}
	\setlength{\tabcolsep}{0.1pt}
	\begin{tabular*}{\hsize}{@{\extracolsep{\fill}}lcccc}
		\toprule
		& \multicolumn{4}{c}{Visible Segmentation (ADE20K)} \\
		\cmidrule(r){2-5} 
		Method           & Modality & mIoU  & mAcc      & aAcc    \\ \midrule
		Deeplab V3+      & VIS  & 44.99 & 55.81      & 81.35 \\
		PSPNet           & VIS  & 44.39 & 54.74      & 81.09 \\
		UPerNet          & VIS  & 43.82 & 54.74      & 81.02 \\
		SegFormer        & VIS  & 49.62 & 61.46      & 83.00 \\ \midrule
		C-JEPA           & VIS  & 38.68 & 48.86      & 79.07 \\
		GATE             & VIS  & 40.51 & 54.90      & 79.68 \\
		PE Rescale       & VIS  & 43.27 & 54.36      & 80.74 \\
		CR2PQ            & VIS  & 47.00 & 57.40      & 83.10 \\  
		ViM-VQ           & VIS  & 48.80 & 59.20      & 82.40 \\   \midrule
		MCMAE            & VIS  & 50.54  & 62.00     & 84.06   \\
		INFMAE           & VIS  & 36.86  & 46.77     & 78.49   \\
		UNIV$^\ast$      & VIS  & 49.93  & 61.66     & 84.00   \\
		UNIV$^\dagger$   & VIS  & 49.81  & 61.74     & 83.84   \\
		IAS (Image-style) & VIS  & 50.60  & 62.24     & \textbf{84.25}  \\
		IAS (UNIV-style)  & VIS  & \textbf{50.82}  & \textbf{62.51}     & 84.15  \\
		\multicolumn{2}{l}{IAS (UNIV-style) vs. UNIV$^\ast$}   & \textcolor{MyForestGreen}{+0.89}  & \textcolor{MyForestGreen}{+0.85}  & \textcolor{MyForestGreen}{+0.15}  \\
		\multicolumn{2}{l}{IAS (UNIV-style) vs. UNIV$^\dagger$}   & \textcolor{MyForestGreen}{+1.01}  & \textcolor{MyForestGreen}{+0.77}  & \textcolor{MyForestGreen}{+0.32}  \\
		\bottomrule
	\end{tabular*}
	\label{tab_ade20k}
\end{table}

\noindent\textbf{Results on Infrared Semantic Segmentation.}
To evaluate the transferability of the proposed IAS framework, we conduct infrared semantic segmentation experiments on four benchmarks. Table~\ref{table_segmentaton} reports comparisons with representative segmentation methods and well-pretrained infrared backbones. For a fair comparison, all IAS variants follow the same MVIP pretraining setting and use the same downstream segmentation architecture. As shown in Table~\ref{table_segmentaton}, IAS consistently outperforms strong pretrained baselines, demonstrating that importance-aware sampling during visible-infrared alignment can learn more transferable infrared representations.

\begin{figure*}
	\centering
	\includegraphics[width=6.8in]{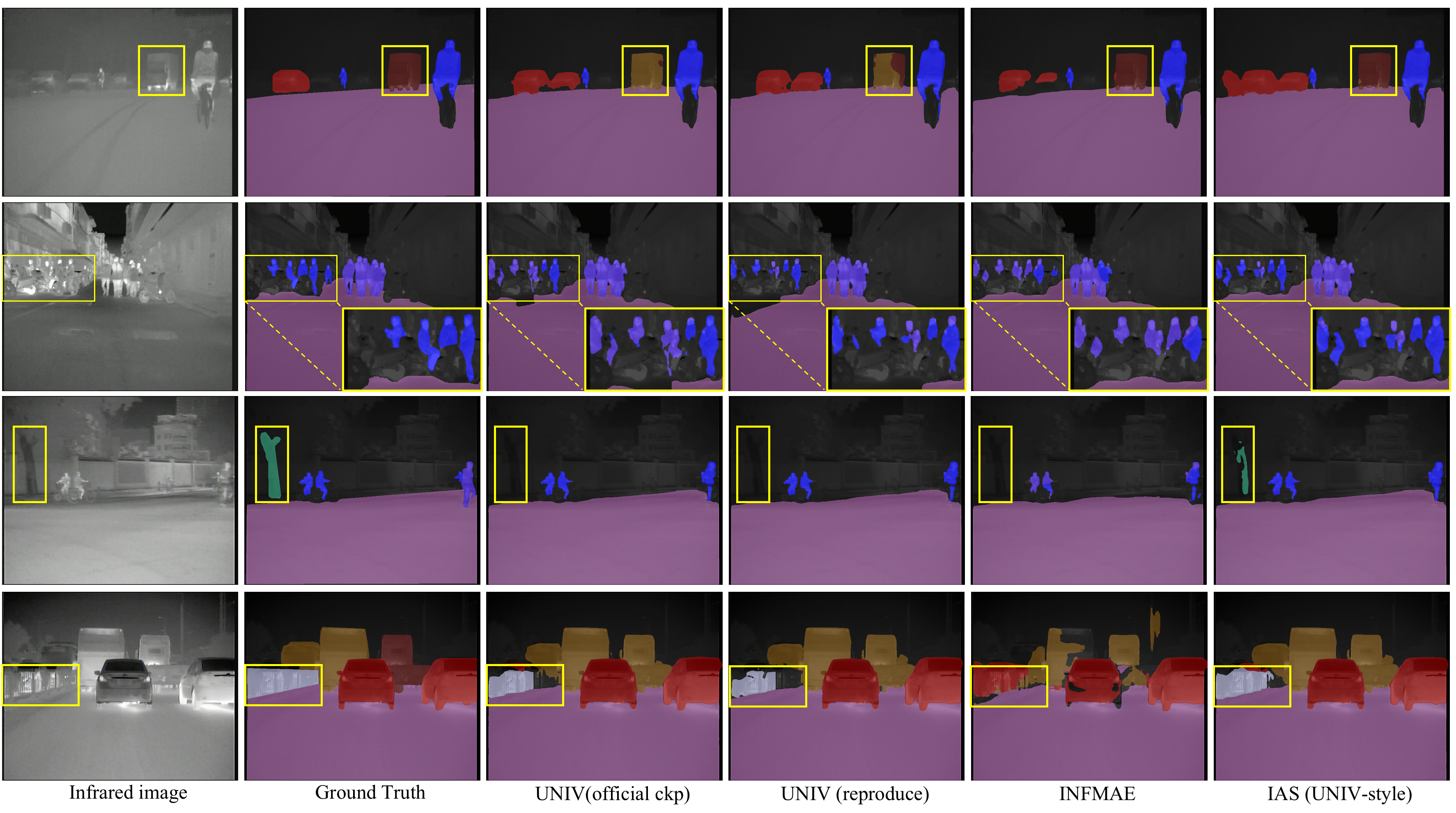}
	\caption{The infrared semantic segmentation experiment result on SCUTSEG dataset.}
	\label{fig_segmentaton_result_scut}
\end{figure*}

Specifically, IAS (UNIV-style) achieves the best mIoU on MFNet, SCUTSEG and MSRS, reaching 52.32\%, 72.23\%, and 75.85\%, respectively. Compared with our reproduced UNIV$^{\dagger}$ baseline, it brings improvements of 1.18, 2.58, and 0.37 points. Meanwhile, IAS (Image-style) obtains the highest mIoU on SODA , achieving 71.07\%, with gains of 1.39 points over UNIV$^{\dagger}$. Although there is a small performance gap between the official UNIV$^{\ast}$ checkpoint and our reproduced UNIV$^{\dagger}$ due to implementation details, IAS consistently improves upon both baselines. Moreover, the experiment results on MFNet and M3FD are shown in Fig.~\ref{fig_segmentaton_result_mfnet} and Fig~\ref{fig_segmentaton_result_scut}, respectively. These results indicate that the proposed sampling strategy is robust and can be seamlessly incorporated into existing visible-infrared pretraining pipelines without modifying downstream segmentation architectures.

\begin{figure*}
	\centering
	\includegraphics[width=6.8in]{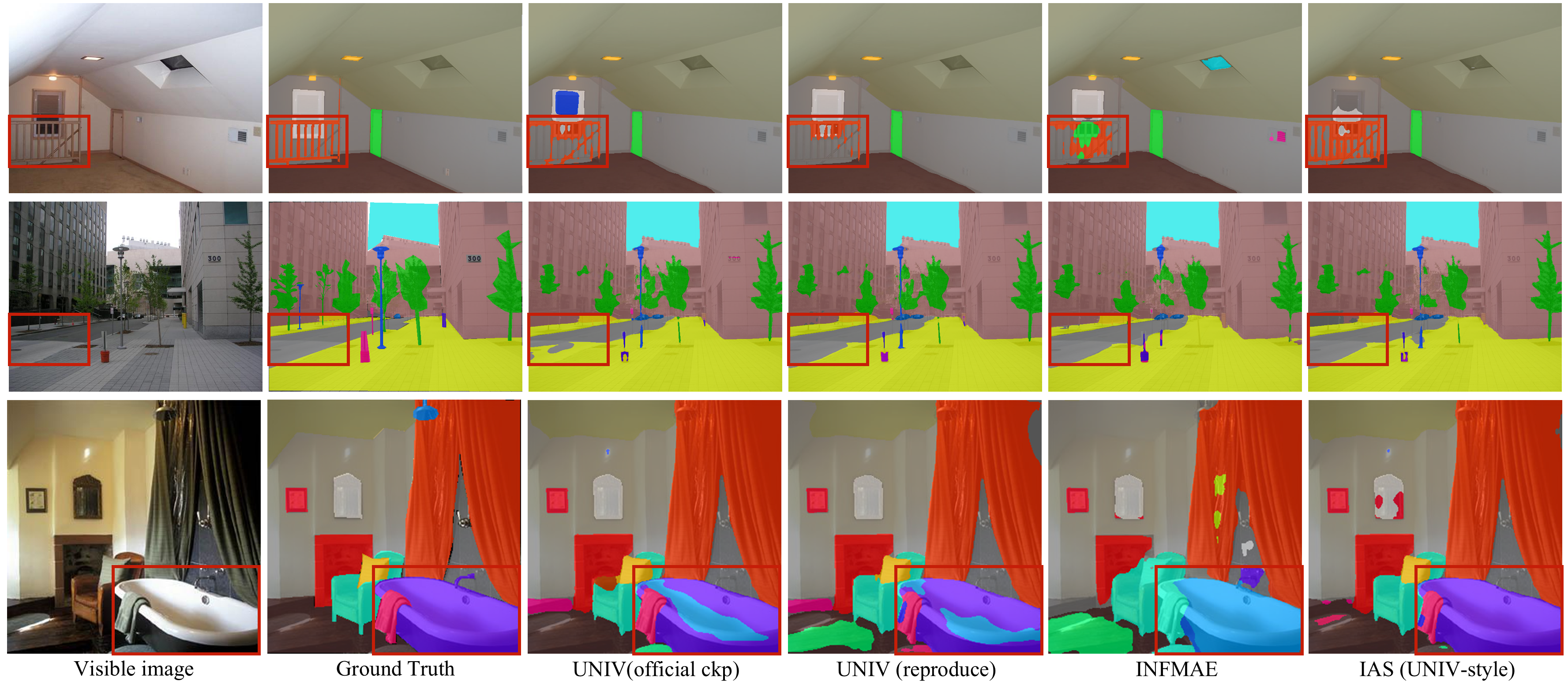}
	\caption{The visible semantic segmentation experiment result on ADE20K dataset.}
	\label{fig_segmentaton_result_ade20k}
\end{figure*}

\noindent\textbf{Results on Visible-Infrared Retrieval.}
To evaluate the quality of the visible-infrared representations learned by our IAS framework, we conduct experiments on the Visible-Infrared retrieval task using the sampled M3FD dataset. Table~\ref{table_retrieval} reports bidirectional retrieval results and compares representative infrared foundation models with our IAS variants under the same pretraining setting. Overall, IAS (Image-style) achieves the best performance and shows large, consistent improvements in both directions, reaching 56.71\% R@1 for visible2infrared and 58.07\% R@1 for infrared2visible, together with substantial gains on R@5 and R@10. In contrast, MAE-based baselines such as MCMAE, INFMAE, and UNIP perform poorly, which suggests that reconstruction-oriented pretraining alone is insufficient for discriminative cross-modal retrieval. Moreover, although a gap is observed between UNIV from the official checkpoint (\(^{\ast}\)) and our reproduction (\(^{\dagger}\)) due to training details, IAS (UNIV-style) still yields consistent gains over UNIV\(^{\dagger}\) on visible2infrared, indicating that IAS can strengthen transferable visible-infrared representations.

\noindent\textbf{Results on Infrared Object Detection.}
To evaluate the capacity of infrared feature representation, we conduct object detection experiments on the M3FD dataset. As reported in Table~\ref{tab_m3fd}, IAS (UNIV-style) achieves the best mAP and AP75, reaching 56.17\% and 60.83\%, respectively. Compared with the official UNIV$^{\ast}$ baseline, it brings gains of 1.00 and 2.18 points. IAS (Image-style) obtains the highest AP50 of 88.33\%, indicating stronger localization ability under a relatively loose IoU threshold. Although IAS uses only infrared images, it achieves competitive or superior performance compared with visible-infrared fusion-based methods. Moreover, the experiment results are shown in Fig~\ref{fig_detection_result_m3fd}. These results suggest that ISA improves infrared representation learning and benefits downstream detection without changing the detection architecture, further demonstrating the transferability of the proposed IAS framework.
\begin{table}[t]
	\centering
	\setlength{\tabcolsep}{2pt}
    \renewcommand{\arraystretch}{1.05}
	\caption{Illustration of IAS generalization across different alignment paradigms.}
	\begin{tabular*}{\hsize}{@{\extracolsep{\fill}}lcccc}
		\toprule
		Pre-training framework & MFNet & SODA   & SCUTSEG  & VIR \\
		\midrule
		Image-style w/o IAS     & 50.06 & 70.40  & 70.26   & 91.78 \\
		Image-style with IAS    & 50.86 & 71.07  & 71.02   & 92.12 \\
		with vs. w/o IAS        & \textcolor{MyForestGreen}{+0.80} & \textcolor{MyForestGreen}{+0.67} & \textcolor{MyForestGreen}{+0.76} & \textcolor{MyForestGreen}{+0.34} \\
		\midrule
		Patch-style w/o IAS     & 15.46 & 45.54  & 53.05 & 1.05 \\
		Patch-style with IAS    & 17.10 & 48.74  & 55.70 & 1.03  \\
		with vs. w/o IAS        & \textcolor{MyForestGreen}{+1.64} & \textcolor{MyForestGreen}{+3.20}  & \textcolor{MyForestGreen}{+2.65} & -0.02 \\
		\midrule
		UNIV-style w/o IAS      & 51.14 & 69.68  & 69.65 & 24.94  \\
		UNIV-style with IAS     & 52.32 & 70.59  & 72.23 & 25.83\\
		with vs. w/o IAS        & \textcolor{MyForestGreen}{+1.18} & \textcolor{MyForestGreen}{+0.91} & \textcolor{MyForestGreen}{+2.58} & \textcolor{MyForestGreen}{+0.89} \\
		\bottomrule
	\end{tabular*}
	\label{tab_abla}
\end{table}

\noindent\textbf{Results on Visible Semantic Segmentation.}
To evaluate the cross-scene generalization ability of IAS, we conduct visible semantic segmentation experiments on ADE20K. As shown in Table~\ref{tab_ade20k}, IAS (UNIV-style) achieves the best mIoU and mAcc, reaching 50.82\% and 62.51\%, with gains of 0.89 and 0.85 points over the official UNIV$^{\ast}$ baseline. IAS (Image-style) obtains the highest aAcc of 84.25\%, showing strong overall pixel-level accuracy.  Moreover, the experiment results are shown in Fig~\ref{fig_segmentaton_result_ade20k}. These results indicate that the IAS framework enhances representation learning while maintaining strong transferability to visible scenes, further validating the generality of the proposed IAS framework.
\begin{table}[t]
	\centering
	\setlength{\tabcolsep}{5pt}
    \renewcommand{\arraystretch}{1.05}
	\caption{Comparison of different importance sampling sources.}
	\begin{tabular*}{\hsize}{@{\extracolsep{\fill}}lccc}
		\toprule
		Importance source & MFNet & SODA & SCUTSEG \\
		\midrule
		None (uniform)             & 51.14 & 69.68 & 69.65 \\
		Sobel gradient             & 52.14 & 70.17 & 70.63 \\
		HOG                        & 52.13 & 69.92 & \textbf{71.71} \\
		Learned with Sobel warm-up & \textbf{52.16} & \textbf{70.50}  & 71.04 \\
		\bottomrule
	\end{tabular*}
	\label{tab:sources}
\end{table}
\begin{table}[t]
	\centering
    \renewcommand{\arraystretch}{1.05}
	\setlength{\tabcolsep}{5pt}
	\caption{Comparison of hard and soft sampling on downstream infrared semantic segmentation datasets}
	\begin{tabular*}{\hsize}{@{\extracolsep{\fill}}lccc}
		\toprule
		Strategy & MFNet & SODA  & SCUTSEG \\
		\midrule
		Hard top-30\%   & 51.94 & 70.50 & 70.02 \\
		Hard top-50\%   & 51.98 & 70.30 & 70.46  \\
		Soft weights    & \textbf{52.32} & \textbf{70.59}  & \textbf{72.23} \\
		\bottomrule
	\end{tabular*}
	\label{tab:hardsoft}
\end{table}
\begin{table}[t]
	\centering
    \renewcommand{\arraystretch}{1.05}
	\caption{Result on patch curriculum schedules with IAS framework.}
	\begin{tabular*}{\hsize}{@{\extracolsep{\fill}}lccc}
		\toprule
		Curriculum & MFNet  & SODA & SCUTSEG\\
		\midrule
		No curriculum & 51.16  & 70.50  & 71.04\\
		Cosine 20\% $\rightarrow$ 100\%  & 51.93  & 69.96 & 70.59 \\
		Linear 20\% $\rightarrow$ 100\%  & \textbf{52.32}  & \textbf{70.59} & \textbf{72.23} \\
		\bottomrule
	\end{tabular*}
	\label{tab:curriculum}
\end{table}
\subsection{Ablation Studies}
\noindent\textbf{Effectiveness and Generalization of IAS.}
We integrate IAS into three representative alignment schemes and evaluate its transferability on infrared semantic segmentation and Visible-Infrared retrieval. As shown in Table~\ref{tab_abla}, IAS consistently improves performance across schemes and tasks, demonstrating strong generalization. For the image-style scheme, IAS yields stable improvements on all datasets, improving MFNet by 0.80, SODA by 0.67, SCUTSEG by 0.76, and VIR by 0.34. For the patch-style scheme, IAS produces larger gains on segmentation, improving SODA by 3.20 and SCUTSEG by 2.65, while VIR remains essentially unchanged, which suggests that patch-only alignment is less suitable for retrieval in this setting. For the UNIV-style scheme, IAS improves both segmentation and retrieval, improving MFNet by 1.18, SODA by 0.91, SCUTSEG by 2.58, and VIR by 0.89. These results highlight that IAS is a robust and plug-and-play component that strengthens visible-infrared representations across different pre-training pipelines.

\noindent \textbf{Impact of Importance Sampling Sources.}
Table~\ref{tab:sources} compares different sources of importance weights. Both hand-crafted infrared structural priors provide strong gains over uniform weighting, where Sobel improves MFNet from 51.14\% to 52.14\%  and SODA from 69.68\%  to 70.17\%, while HOG achieves the best mIoU at 71.71\%  on SCUTSEG. The learned importance module with Sobel warm-up performs best overall, reaching 52.16\%  on MFNet and 70.50 on SODA, which supports that learning importance is most effective when initialized with a simple structural prior.

\begin{figure}
	\centering
	\includegraphics[width=3.2in]{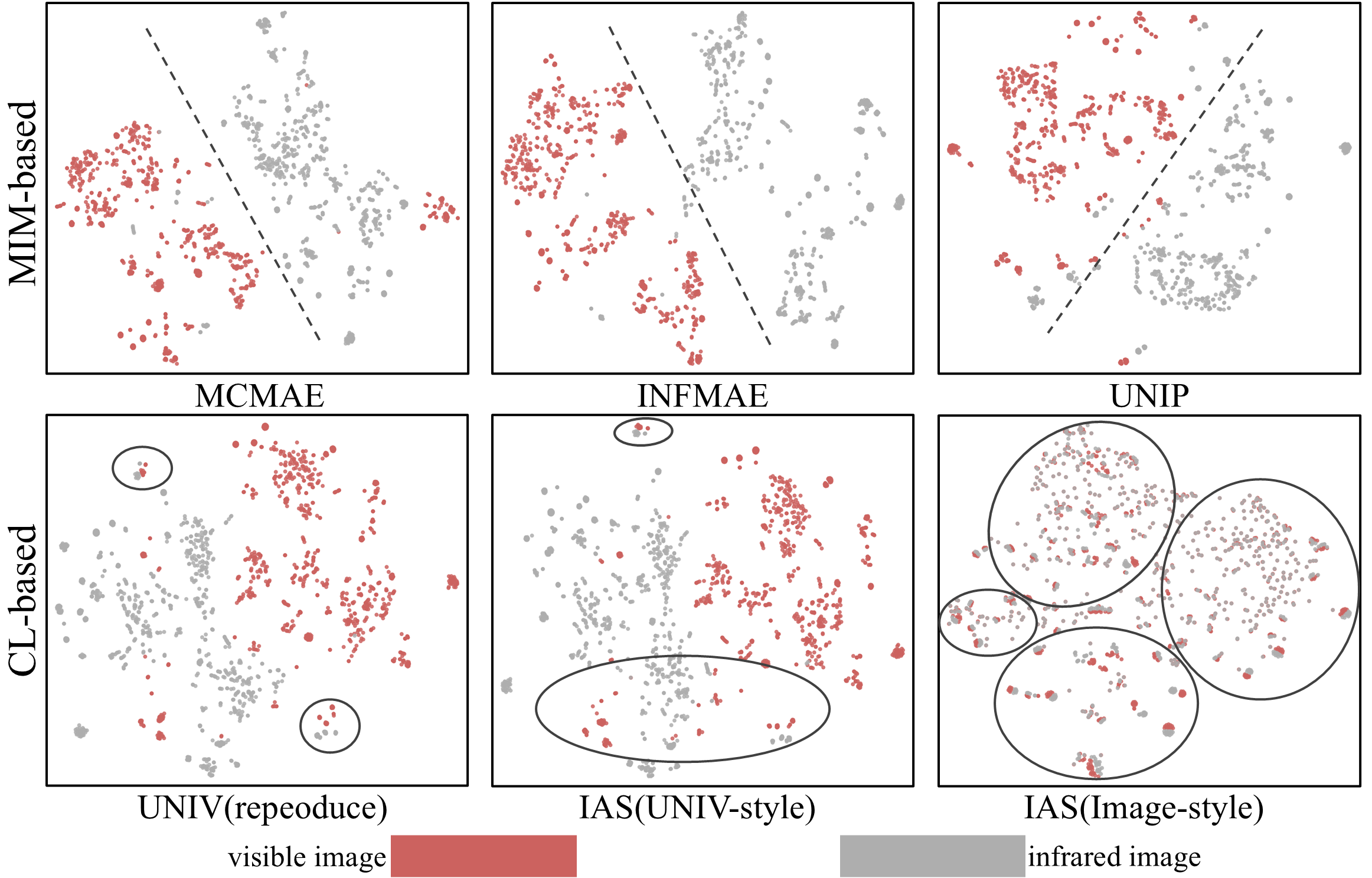}
	\caption{Visualization of feature on visible-infrared retrieval dataset.}
	\label{fig_retrieval_feature}
\end{figure}

\noindent\textbf{Hard Truncation vs.\ Soft Weighting.}
Table~\ref{tab:hardsoft} compares hard truncation and soft weighting on infrared semantic segmentation. Soft weighting consistently performs best on mIoU metrics, and we attribute this advantage to the fact that soft weighting preserves informative gradients from all patches while suppressing unreliable ones, which leads to smoother optimization and better performance.

\noindent\textbf{Effect of Patch Curriculum.}
Table~\ref{tab:curriculum} compares different curriculum schedules on IAS(UNIV-style) with learnable sampling source. The linear schedule yields the best overall performance across datasets, demonstrating that linear schedule provides the best trade-off between convergence speed and final performance.
\begin{table*}[t!]
	\centering
	\renewcommand{\arraystretch}{1.05}
	\caption{Settings of downstream segmentation and detection task.}
	\label{tab_downstream_settings}
	\setlength{\tabcolsep}{1pt}
	\begin{tabular*}{\hsize}{@{\extracolsep{\fill}}lcccccc}
		\toprule
		Hyperparameters        & SODA & MFNet & SCUTSEG & MSRS & ADE20K & M3FD\\
		\midrule
		Input resolution       & $512 \times 512$ & $512 \times 512$ & $512 \times 512$ & $512 \times 512$ & $512 \times 512$ & $1024 \times 1024$ \\
		Pre-training epochs    & $400$        & $400$   & $400$   & $400$   & $400$   & $400$  \\
		Fine-tuning iterations & $14400$      & $19600$ & $16800$ & $48000$ & $160000$ & $168000$\\
		Learning rate     & $5\mathrm{e}{-4}$ & $5\mathrm{e}{-4}$ & $5\mathrm{e}{-4}$ & $1\mathrm{e}{-4}$ & $5\mathrm{e}{-4}$ & $8\mathrm{e}{-5}$ \\
		Batch size             & $8$          & $8$     & $8$     & $8$       & $8$          & $2$  \\
		Optimizer              & AdamW        & AdamW   & AdamW   & AdamW     & AdamW        & AdamW \\
		Weight decay           & $0.05$       & $0.05$        & $0.05$        & $0.05$       & $0.05$       & $0.1$ \\
		Optimizer momentum     & $\beta_1,\beta_2=0.9,0.999$  & $\beta_1,\beta_2=0.9,0.999$ & $\beta_1,\beta_2=0.9,0.999$  & $\beta_1,\beta_2=0.9,0.999$ & $\beta_1,\beta_2=0.9,0.999$  & $\beta_1,\beta_2=0.9,0.999$ \\
		Learning rate schedule & Poly decay   & Poly decay    & Poly decay    & Poly decay   & Poly decay & Cosine decay   \\
		Minimal learning rate  & $0$          & $0$           & $0$           & $0$          & $0$       & $0$\\
		Warmup steps           & $1500$       & $1000$        & $1700$        & $1500$       & $1500$    & $210$ \\
		\bottomrule
	\end{tabular*}
\end{table*}

\noindent\textbf{Quantitative overhead analysis.}
We analyze the additional cost introduced by the proposed IAS framework. IAS keeps the backbone unchanged and predicts patch importance only from infrared patch tokens. Specifically, the Sobel, HOG, and Random variants introduce no learnable parameters. The learnable variant adds only 0.107M parameters and 0.0418 GFLOPs per image, which is marginal compared with the backbone. This analysis demonstrates that IAS is a low-cost and plug-and-play module.
\begin{figure}
	\centering
	\includegraphics[width=3.2in]{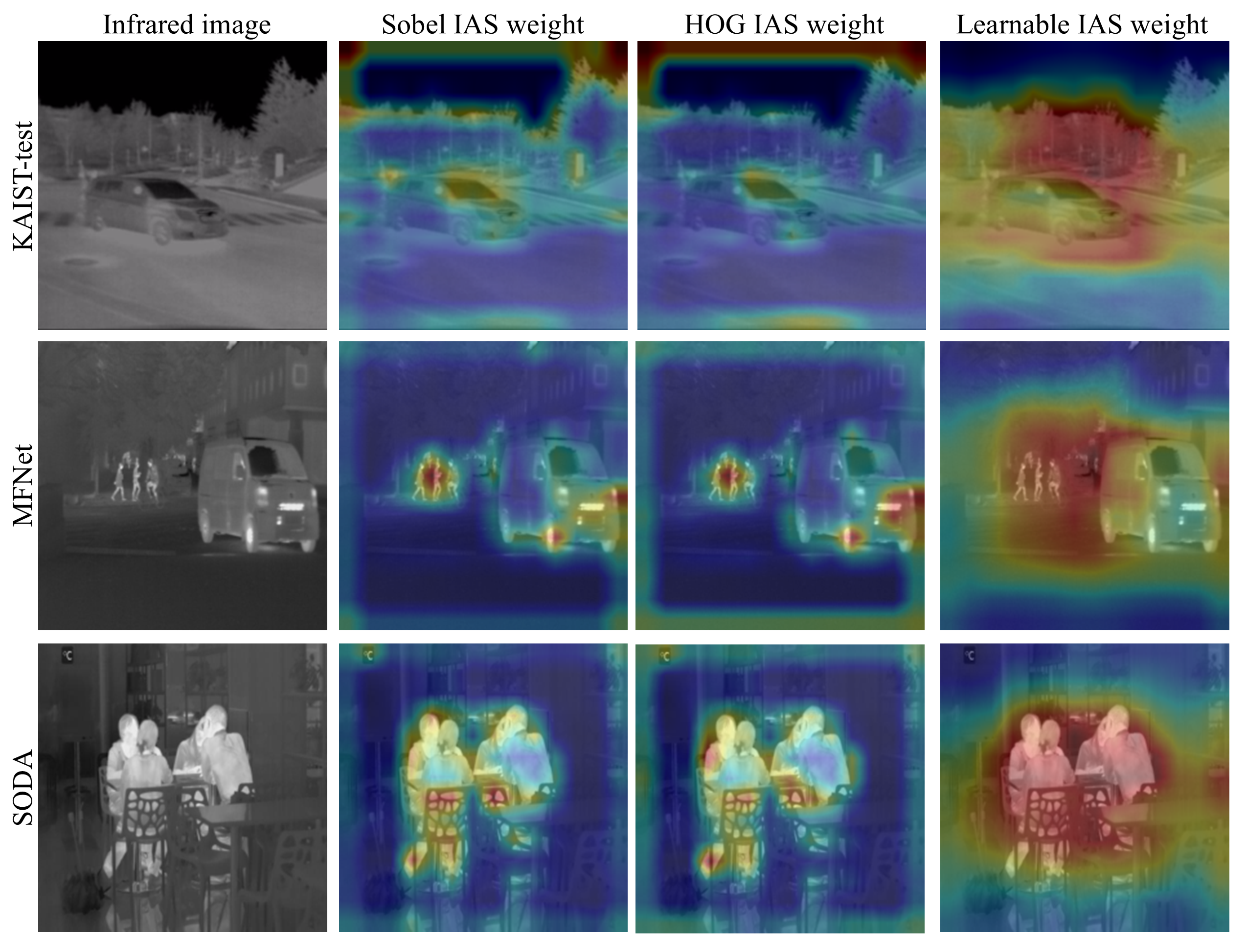}
	\caption{Visualization of heatmaps of IAS weight.}
	\label{fig_ias_vis}
\end{figure}

\section{Visualization}
To demonstrate the effectiveness of IAS, we provide visualizations\textcolor{red}{\footnote{The visualized samples in this section are not included in the pre-training dataset.}} of our framework on several cases as follows: 

\noindent\textbf{Feature visualization.} We use t-SNE~\cite{t-SNE} to visible-infrared features extracted by the proposed IAS. As shown in Fig.~\ref{fig_retrieval_feature}, it can be observed that the visible and infrared features of the same image in our framework achieve a much closer distribution. This indicates that the proposed IAS yields superior visible-infrared joint representation.

\begin{figure}
	\centering
	\includegraphics[width=3.2in]{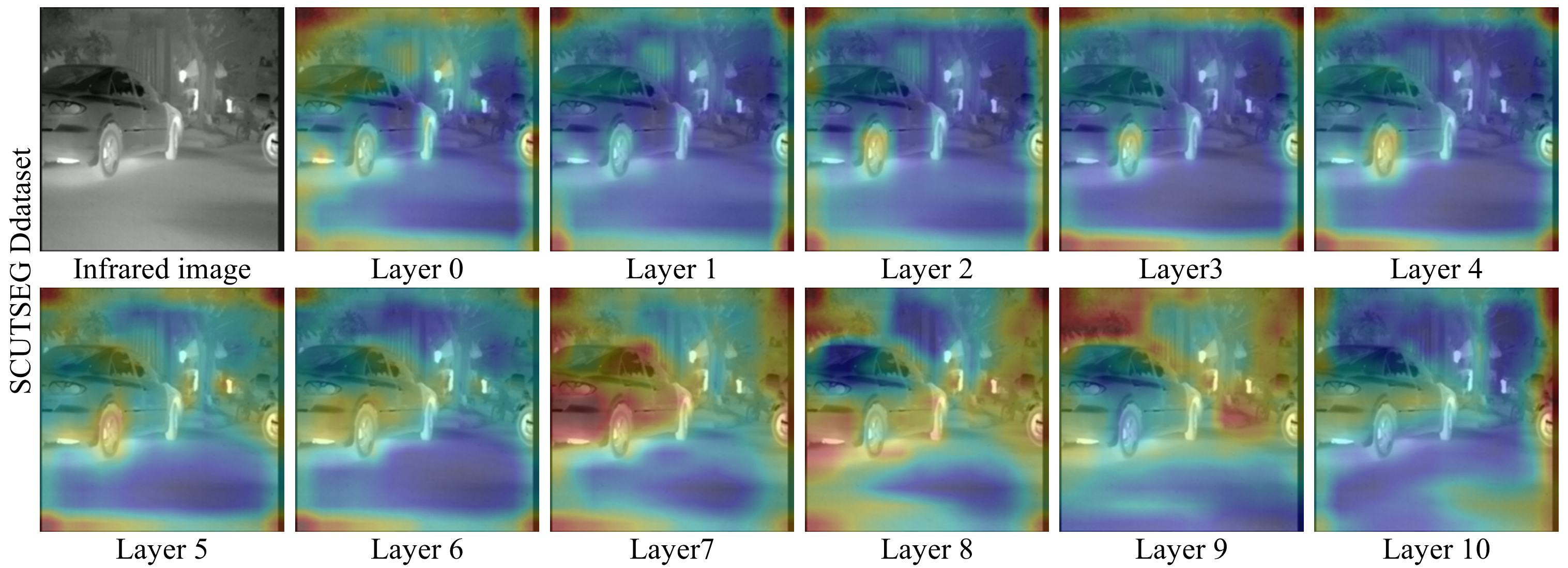}
	\caption{The visualization of layers of our IAS backbone.}
	\label{fig_vis_layer}
\end{figure}
\begin{figure}
	\centering
	\includegraphics[width=3.2in]{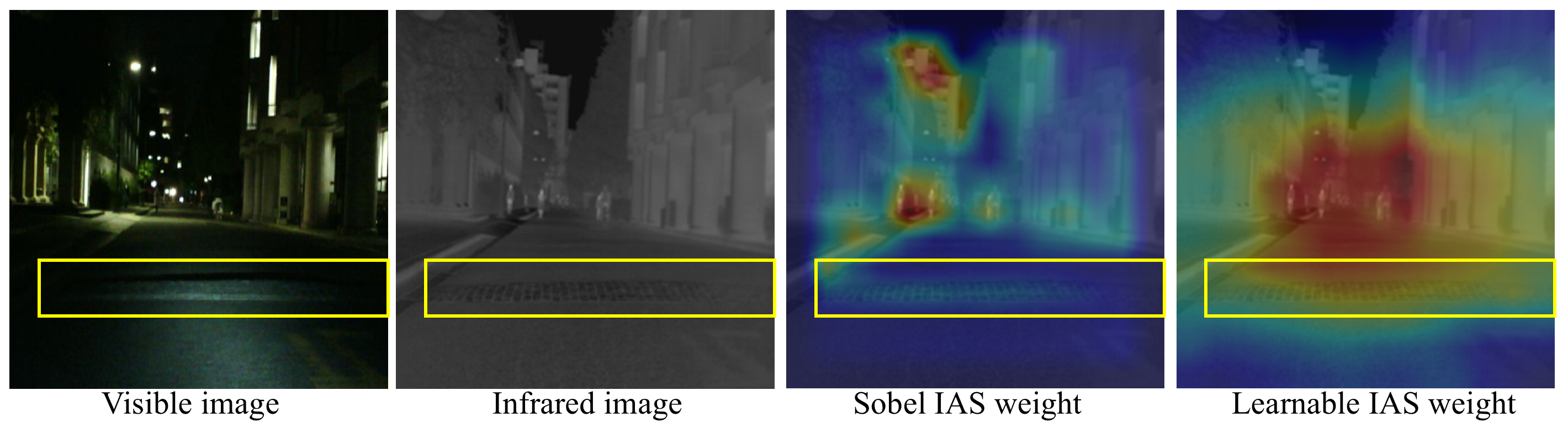}
	\caption{The visualization of special region issue(speed bumps).}
	\label{fig_change}
\end{figure}

\noindent\textbf{IAS weight.} We visualize the weight heatmap of three IAS module variants: Sobel, HOG, and learnable weighting schemes. As shown in Fig.~\ref{fig_ias_vis}, it indicates that IAS captures meaningful infrared semantic information and guides the different paradigms of contrastive pretraining. 

\noindent\textbf{Layer visualization.} We visualize the heatmap of different backbone layers. As illustrated in Fig.~\ref{fig_vis_layer}, semantic objects like vehicles and pedestrians exhibit high attention weight, whereas the road surface receives lower weights. This highlights the semantic perception capacity of our method and its transfer capability for downstream tasks.

\section{Case Study}
In this section, we conduct a case study to evaluate the robustness of IAS in infrared-weak regions. Infrared structural cues are informative but may be unreliable in low-contrast areas, such as speed bumps or weak object boundaries. Instead of hard token removal, IAS employs soft importance weighting and a curriculum schedule, enabling low-edge regions to be progressively integrated as training proceeds. As shown in Fig.~\ref{fig_change}, the speed bumps highlighted by the yellow boxes receive low attention under the Sobel-based setting, while obtaining higher attention in the learnable setting. This indicates that the proposed design preserves semantically critical regions while leveraging infrared structure as an effective initialization. These results demonstrate that IAS can adaptively refine patch importance and improve robustness in challenging infrared scenarios.

\section{Conclusion}
We revisit visible-infrared alignment pre-training from the perspective of patch sampling and argue that sampling plays a critical role. In contrast to existing methods that uniformly treat all patches, we demonstrate that many patches are either physically mismatched or uninformative across modalities, and that explicitly modeling their importance leads to more effective alignment. To this end, we propose IAS, a simple and modular framework that integrates infrared structural priors, a learnable sampling module, and a patch curriculum, and is compatible with main alignment models. Extensive experiments show consistent performance gains, and ablation studies confirm that the improvement primarily stems from the sampling strategy rather than architectural modifications. In future work, we will extend it to more downstream tasks and other modality pairs such as visible-SAR.

\printcredits
\section*{Data availability}
Data and checkpoints will be made available on request.

\section*{Declaration of competing interest}
The authors declare that they have no conflict of interest.

\section*{Acknowledgments}
This work was supported in part by the National Natural Science Foundation of China under Grant 62525108 and Grant 62371185, in part by the National Key Research and Development Program of China under Grant 2021 YFA0715203, in part by the Science and Technology Inovation Program of Hunan Province under Grant 2024RC1030 and Grant 2023RC3124, in part by the Project of Yuelushan Center for Industrial Innovation under Grant 2025YCII0202.

\bibliographystyle{cas-model2-names}

\bibliography{cas-refs,IEEEabrv,refs}





\end{document}